\documentclass{article}

\usepackage[a4paper,width=150mm,top=25mm,bottom=25mm]{geometry}

\usepackage[utf8]{inputenc} 
\usepackage[T1]{fontenc}    
\usepackage{hyperref}       
\usepackage{url}            
\usepackage{booktabs}       
\usepackage{amsfonts}       
\usepackage{nicefrac}       
\usepackage{microtype}      
\usepackage{lipsum}
\usepackage{graphicx}
\usepackage{subcaption}
\usepackage{xcolor}
\usepackage{authblk}
\usepackage{dsfont}
\usepackage[T1]{fontenc}
\usepackage{lmodern}

\definecolor{newcolor}{rgb}{.8,.349,.1}
\definecolor{shadecolor}{cmyk}{0,0,0,.1}

\newcommand\redsout{\bgroup\markoverwith{\textcolor{red}{\rule[0.5ex]{2pt}{0.4pt}}}\ULon}

\date{}
\title{Hierarchical Generalized Category Discovery for Brain Tumor Classification in Digital Pathology}

\author[1]{Matthias Perkonigg \footnote{corresponding author: matthias.perkonigg@i-med.ac.at}}
\author[1]{Patrick Rockenschaub}
\author[1]{Georg Göbel}
\author[2]{Adelheid Wöhrer}

\affil[1]{Institute of Clinical Epidemiology, Public Health, Health Economics, Medical Statistics and Informatics, Medical University of Innsbruck, Austria}
\affil[2]{Institute of Neuropathology and Neuromolecular Pathology,  Medical University of Innsbruck, Austria}

\begin{document}

\maketitle              
\begin{abstract}
  Accurate brain tumor classification is critical for intra-operative decision making in neuro-oncological surgery. However, existing approaches are restricted to a fixed set of predefined classes and are therefore unable to capture patterns of tumor types not available during training. Unsupervised learning can extract general-purpose features, but it lacks the ability to incorporate prior knowledge from labelled data, and semi-supervised methods often assume that all potential classes are represented in the labelled data. Generalized Category Discovery (GCD) aims to bridge this gap by categorizing both known and unknown classes within unlabelled data. To reflect the hierarchical structure of brain tumor taxonomies, in this work, we introduce Hierarchical Generalized Category Discovery for Brain Tumor Classification (HGCD-BT), a novel approach that integrates hierarchical clustering with contrastive learning. Our method extends contrastive learning based GCD by incorporating a novel semi-supervised hierarchical clustering loss. We evaluate HGCD-BT on OpenSRH, a dataset of stimulated Raman histology brain tumor images, achieving a +28\% improvement in accuracy over state-of-the-art GCD methods for patch-level classification, particularly in identifying previously unseen tumor categories. Furthermore, we demonstrate the generalizability of HGCD-BT on slide-level classification of hematoxylin and eosin stained whole-slide images from the Digital Brain Tumor Atlas, confirming its utility across imaging modalities.
\end{abstract}

\section{Introduction}
For brain tumor patients, surgical intervention is often a critical component of treatment. During surgery, the intra-operative classification of brain tumors is essential to guide personalized decision-making. Depending on the underlying pathological classification the surgical radicality varies between biopsy and supramaximal resection. However, the large diversity and complexity of brain tumor types make it difficult to assemble representative datasets that span the full spectrum of pathologies for training machine learning models. Current classification methods are typically constrained to a set of pathological patterns defined prior to training of a model. In practice, however, datasets often lack pathological patterns of certain subtypes, the number of different subtypes may be unknown (e.g. when sourced from a large-scale, unlabelled database), or include only partially annotated subsets, typically biased toward the most prevalent tumor types.
Unsupervised and semi-supervised methods can achieve remarkable success in extracting general-purpose feature representations that could be leveraged for a wide-range of downstream tasks, as demonstrated by foundation models in digital pathology (e.g.~\cite{chen_towards_2024,xu_whole-slide_2024,vorontsov_foundation_2024}). However, unsupervised learning approaches are inherently unable to incorporate prior knowledge in the form of labelled data, while semi-supervised methods typically assume that all potential classes are present in the labelled subset of the data. These limitations hinder their applicability to neuro-oncology, where datasets are both incomplete and imbalanced.
Generalized Category Discovery (GCD) addresses these limitations by aiming to categorize unlabelled data through a combination of labelled and unlabelled samples. The unlabelled samples may belong to a known category represented in the labelled set or to novel categories that need to be discovered \cite{vaze_generalized_2022}. Different GCD approaches \cite{vaze_generalized_2022,pu_dynamic_2023,zhang_promptcal_2023} have demonstrated strong performance in image analysis tasks using benchmark datasets of natural images.
When extended to brain tumor classification based on imaging, GCD approaches have the potential to reduce the need for extensive labelling, especially for rare or difficult-to-annotate tumor subtypes. Moreover, by revealing previously unrecognized subgroups, these approaches could contribute to a finer-grained understanding of tumor biology and support precision oncology. Brain tumor classification is inherently hierarchical, as reflected in the WHO classification system \cite{louis_2021_2021}. This hierarchical structure encodes clinically meaningful relationships that directly influence diagnosis and treatment. Designing a GCD approach to leverage hierarchical information offers the opportunity to align computational discovery with underlying biological organization, improving robustness and interpretability.

\paragraph{Contribution}
In this work, we propose a novel approach to Generalized Category Discovery (GCD) tailored for the specific challenges of brain tumor classification. Since brain tumor taxonomies naturally follow a hierarchical structure, therefore the key idea of our approach is to explicitly model this hierarchy to capture the underlying pathological concepts better than existing GCD methods focusing mainly on contrastive learning.
To this end we introduce Hierarchical Generalized Category Discovery for Brain Tumor Classification (HGCD-BT), a method that incorporates this hierarchical organization directly during the training process. This is achieved by extending the unsupervised hierarchical loss proposed by \cite{znalezniak_contrastive_2023} to a novel semi-supervised hierarchical clustering loss function specifically tailored to the setting of GCD.
We conduct experiments on a dataset of Stimulated Raman Histology (SRH) data of brain tumors \cite{jiang_opensrh_2022} demonstrating that HGCD-BT shows promising performance for GCD for brain tumor classification and outperforms GCD methods relying soley on contrastive learning. Furthermore, we validate its generalizability on hematoxylin and eosin (H\&E)–stained whole-slide images (WSIs) \cite{roetzer-pejrimovsky_digital_2022}, achieving accurate slide-level classification across 12 tumor types. These results highlight the robustness of HGCD-BT across both imaging modalities and classification granularities.

\section{Related work}

\paragraph{GCD} is a setting in ML in which the goal is to classify images in a dataset, a subset of which has known labels. GCD aims at assigning labels to all remaining images, using class labels that may or may not have been observed in the labeled subset. The setting of GCD has been primarily explored in natural image classification \cite{vaze_generalized_2022,pu_dynamic_2023,zhang_promptcal_2023}. In addition, GCD has been combined with active learning \cite{ma_active_2024} or continual learning \cite{wu_metagcd_2023,zhao_incremental_2023}. Particularly relevant to our work are GCD methods considering the hierarchical nature of categories. To leverage this inherent hierarchy, a variety of techniques have been proposed. These include self-coding to implicitly learn a category tree \cite{rastegar_learn_2023}, hierarchical pseudo-labeling \cite{rastegar_selex_2024}, semi-supervised hierarchical clustering \cite{hao_cipr_2024} or hierarchical prototyping \cite{wang_discover_2023}. \cite{fan_seeing_2024, liu_debgcd_2025} use the properties of the hyperbolic space to model hierarchical relationships and improve category discovery. GCD is rarely applied to medical imaging, one approach \cite{fan_seeing_2024} explores discovery of concepts via probabilistic modeling in diverse medical imaging datasets.

\paragraph{Artificial Intelligence for Stimulated Raman Histology}
SRH, a label-free optical imaging technique, able to produce virtual histology images in 2-3 minutes was introduced in the clinic \cite{orringer_rapid_2017, hollon_near_2020} and has the potential to revolutionize intraoperative diagnosis. SRH utilizes laser-based imaging to identify variations in macromolecule concentration within biomedical specimens, and creates imaging contrast based on these differences. It has been demonstrated that SRH can be used to provide non-inferior human diagnosis compared to frozen section diagnosis in a significantly shorter timeframe \cite{einstein_stimulated_2022}.
The inherently digital nature of SRH images and their real-time application during surgery led to the development of various ML methods, particularly for the classification of different brain tumor types, primarily covering the more common tumor types. Convolutional Neural Networks (CNNs) were used in fully-supervised algorithms to classify SRH images in low quality, tumor and non-tumor regions \cite{reinecke_novel_2022}, to differentiate between patches of tumor recurrence and pseudoprogression \cite{hollon_rapid_2021} or to classify 13 diagnostic tumor types \cite{hollon_rapid_2021}. Self-supervised representation learning was applied to the differentiation between primary Central nervous system (CNS) lymphoma (PCNSL) and other CNS entities \cite{reinecke_fast_2024}.

\paragraph{Artificial Intelligence for FFPE brain tumor classification}
Artificial Intelligence has been extensivley applied to hematoxylin and eosin (H\&E)-stained slides of formalin-fixed and paraffin-embedded (FFPE) for brain tumor subtyping. Specifically, \cite{redlich_applications_2024} identified 23 studies dealing with glioma subtyping. Those studies focus on classifying a limited set of two to five common subtypes. Besides studies specifically build for brain tumor classification, different foundation models in pathology \cite{chen_towards_2024, wang_pathology_2024} use brain tumor subtyping as a downstream task to demonstrate the model. They demonstrate promising performance by applying linear probing on tasks such as predicting IDH status \cite{chen_towards_2024, wang_pathology_2024, xiang_visionlanguage_2025} or common tumor subtyping \cite{chen_towards_2024}. However, different from GCD methods, linear probing requires annotation for all tumor subtypes.

\section{Method}
\begin{figure}[h]
  \centering
  \includegraphics[width=0.95\linewidth]{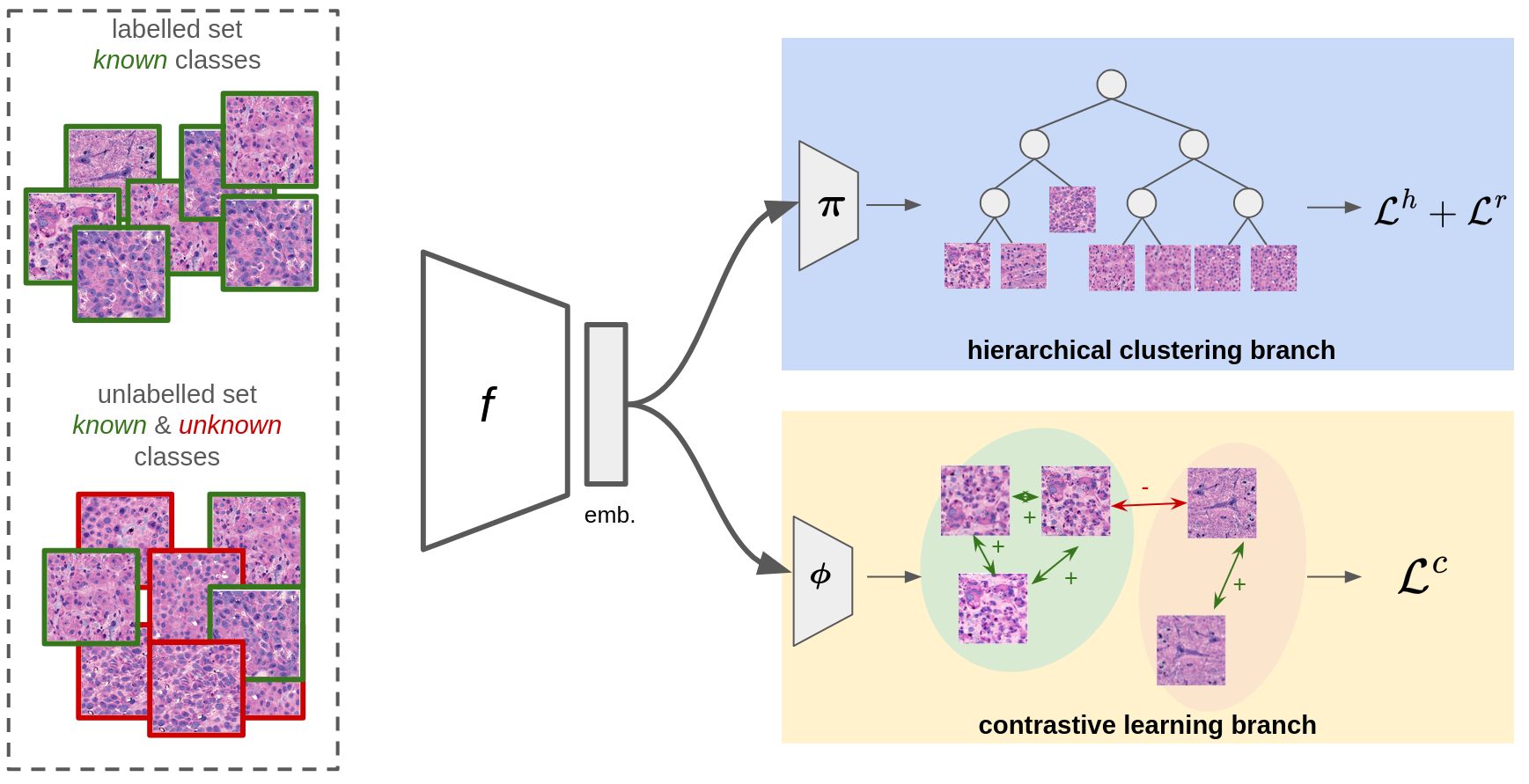}
  \caption{Overview of our proposed method HCGD-BT consisting of a hierarchical clustering branch and a contrastive learning branch.}
  \label{fig:overview}
\end{figure}

GCD is a framework for classifying images when only a subset of the dataset has known class labels. The core goal is to assign appropriate labels to the remaining unlabelled images, which may belong to both known and previously unseen classes. Formally, let $\mathcal{X}$ denote the input space and $\mathcal{Y}$ the label space. The dataset $\mathcal{D} = \mathcal{D}_l \cup \mathcal{D}_u$ consists of a labelled subset $\mathcal{D}_l = \{x_i, y_i\}^N_{i=0} \in \mathcal{X} \times \mathcal{Y_L}$ and an unlabelled subset $\mathcal{D}_u = \{x_i, y_i\}^M_{i=0} \in \mathcal{X} \times \mathcal{Y_U}$, where the label space of the labelled dataset is a subset of the label space of the unlabelled dataset $\mathcal{Y_L} \subset \mathcal{Y_U}$. During training the labels in $\mathcal{Y_U}$ are not accessible. After training labels should be estimated for every sample in $\mathcal{D}_u$.

Figure~\ref{fig:overview} illustrates the architecture of our proposed method. HGCD-BT utilizes a (pre-trained) feature encoder $f$, and two separate branches, namely a hierarchical clustering branch using a novel loss formulation (see Section \ref{sec:hierarchical_clustering}) and a contrastive learning branch (see Section \ref{sec:contrastive_learning}). The feature encoder and the two branches are trained jointly to achieve category discovery. The contrastive learning branch structures the feature space such that data points of the same class are close together. The hierarchical learning branch enforces a hierarchy which clusters data points by a sequence of decisions.

\subsection{Contrastive learning loss}
\label{sec:contrastive_learning}
For the contrastive loss we follow the concept proposed in \cite{vaze_generalized_2022}, which combines an unsupervised contrastive loss and a supervised contrastive loss to utilize both the labelled $\mathcal{D}_l$ and unlabelled $\mathcal{D}_u$ data.
The unsupervised loss \cite{chen_simple_2020} is computed from two randomly augmented views $x_i$ and $x_i'$ of the same input in a mini-batch $\mathcal{B}$. First, both views are embedded by using the feature extractor $f$ followed by a projection head $\phi$ into an embedding vector $\mathbf{z}_i = \phi(f(x_i))$. Then, the loss is given as:
\begin{equation}
  \mathcal{L}^u_i =   -\log \frac{\exp(\mathbf{z}_i \cdot \mathbf{z}_i')}{\sum_{n=1}^{|\mathcal{B}|} \mathds{1}_{[n\neq i]}\exp(\mathbf{z}_i \cdot \mathbf{z}_n)},
\end{equation}
where $\mathds{1}_{[n\neq i]}$ is the indicator function, equal to 1 when $n \neq i$ and 0 otherwise.
For the supervised loss we define $\mathcal{P}(i)$ as the indices of those images with the same label as $x_i$ in $\mathcal{B}$. Furthermore, $\mathcal{B}_l$ denotes the labeled subset within the mini-batch $\mathcal{B}$.
We can then write the supervised loss as:
\begin{equation}
  \mathcal{L}^s_i =   -\frac{1}{|\mathcal{P}(i)|} \sum_{q\in \mathcal{P}(i)} \log \frac{\exp(\mathbf{z}_i \cdot \mathbf{z}_q)}{\sum_{n=1}^{|\mathcal{B}_l|}  \mathds{1}_{[n\neq i]}\exp(\mathbf{z}_i \cdot \mathbf{z}_n)}
\end{equation}
Finally the loss for a whole mini-batch is given as:
\begin{equation}
  \mathcal{L}^c = (1-\alpha) \sum_{i\in \mathcal{B}} \mathcal{L}^u_i + \alpha \sum_{j\in \mathcal{B}_l} \mathcal{L}^s_j
  \label{eq:contrastive_loss}
\end{equation}
with a weight coefficient $\alpha$ to balance the unsupervised and supervised loss.

\subsection{Semi-supervised hierarchical clustering loss}
\label{sec:hierarchical_clustering}
To reflect the hierarchical structure of brain tumor diagnostics, inspired by CoHiClust \cite{znalezniak_contrastive_2023} we propose to use contrastive hierarchical clustering in the training phase of the model to design a loss that takes the hierarchical nature of pathological patterns into account. To utilize the prior information in form of the labelled dataset we extend the unsupervised loss in \cite{znalezniak_contrastive_2023} to a semi-supervised formulation.
To formulate the hierarchical tree loss we follow \cite{znalezniak_contrastive_2023} to construct a soft binary decision tree. Different from hard decision trees, within a soft decision tree every internal node defines a probability of taking the right or left branch. The idea is that similar data points will be routed through the same nodes of the tree. We parametrize this tree by adding a projection head $\pi$ after the encoder $f$. $\pi$ maps the features extracted by $f$ to a $K$-dimensional vector where $K=2^T-1$ with $T$ being the height of the binary decision tree. The parameter $K$ denotes the number of internal nodes in the tree.
The projection head $\pi$ is modeled as a simple feed-forward network that uses the sigmoid function $\sigma$ to generate outputs that can be interpreted as the probabilities of taking the left branch of an internal node.
Based on these calculated probabilities and the observation that similar data points should follow the same path through the tree, we can formulate a loss function.

Formally, for two inputs $x_1, x_2$ we can consider their posterior probabilities  $P_t(x_1), P_t(x_2)$ at tree level $t$ and compute the probability that they reach the same node at level $t$ as the scalar product $P_t(x_1) \cdot P_t(x_2)$, to avoid trivial solutions \cite{znalezniak_contrastive_2023} we further formulate this probability using the Bhattacharyya coefficient \cite{bhattacharyya_measure_1946}:
\begin{equation}
  s_t(x_1, x_2) = \sqrt{P_t(x_1) \cdot P_t(x_2)} = \sum_{i=0}^{2^t-1}\sqrt{P_t^i(x_1)P_t^i(x_2)}
\end{equation}
For a final similarity function over the whole tree those level-wise similarities are summed: $s(x_1, x_2) = \sum_{t=0}^{T-1}s_t(x_1, x_2)$.

For an input $x_i$ and its augmented version $x_i'$ in a batch $B$ our contrastive hierarchical loss is written as:
\begin{equation}
  \mathcal{L}^h_i = \frac{1}{|\mathcal{N}(i)|} \sum_{j\in \mathcal{N}(i)} s(x_i, x_j) - \frac{1}{|\mathcal{P}(i)|+1} \left[\left(\sum_{j\in \mathcal{P}(i)} s(x_i, x_j)\right)+s(x_i, x_i')\right]
\end{equation}
where $\mathcal{N}(i)$ are the indices of inputs with different labels than $x_i$, and $\mathcal{P}(i)$ are those that share a label with $x_i$.
Note that, if $x_i \in B_l$, $\mathcal{N}(i)$ contains all samples in $B$ that have a different label than $x_i$ and those for which no label is provided. If $x_i \notin B_l$ $\mathcal{N}(i)$ contains all samples from $B$ except $x_i$.

To avoid solutions where only parts of the tree are used for clustering, we employ the regularization loss $\mathcal{L}^r$ proposed in \cite{znalezniak_contrastive_2023}, which encourages the model to use both the left and the right sub-tree. This is achieved by minimizing the cross-entropy between the desired balanced distribution $[0.5, 0.5]$ and the mean probabilities of all samples in $B$ at each internal node.

The overall loss to train our proposed HGCD-BT method is written as:
\begin{equation}
  \mathcal{L} = \mathcal{L}^h + \beta \mathcal{L}^r + \gamma \mathcal{L}^c
  \label{eq:overall_loss}
\end{equation}

where $\beta$ and $\gamma$ are weighting factors to balance the influence of the loss functions.

\section{Experiments and Results}
\label{sec:experiment_setup}
\subsection{Data}
We used two datasets for evaluating the proposed approach, a dataset of Stimulated Raman histology \textit{OpenSRH} \cite{jiang_opensrh_2022} for patch-level classification, and the Digital Brain Tumor Atalas (DBTA) for slide-level classification of H\&E-stained WSIs.

\subsubsection*{OpenSRH}
\textit{OpenSRH} \cite{jiang_opensrh_2022} is a publicly available image classification dataset consisting of SRH data from brain tumor patients. The dataset comprises 1348 SRH slides from 307 patients, annotated with pathological labels for six brain tumor types — high-grade glioma (HGG), low-grade glioma (LGG), metastases, meningioma, schwannoma, pituitary adenoma — along with a normal tissue category. The slides are divided into non-overlapping 300x300 pixel patches resulting in 282.931 patches, which serve as inputs for our category discovery model. For evaluation of the category discovery we designated four classes as known (HGG, meningioma, metastases, normal) and three as unknown or novel classes (LGG, schwannoma, pituitary adenoma). Among the known classes, we set a proportion of 50\% as labelled in the training data. Image processing follows the methodology in \cite{jiang_opensrh_2022} to generate three-channel image patches.

\subsubsection*{Digital Brain Tumor Atlas (DBTA)}
The DBTA dataset \cite{roetzer-pejrimovsky_digital_2022} is an openly available dataset of digitized H\&E- stained brain tumour slides from FFPE tissue. It contains over 126 different brain tumor subtypes. For this study, we selected a subset of the dataset containing tumor types with at least 50 samples each, yielding 12 classes and a total of 1,209 whole-slide images (WSIs). From these we randomly sampled 7 classes as known and the remaining 5 as novel classes. An overview of the selected subset including number of WSIs and split into known and novel classes is provided in Table \ref{tab:dbta}. From this we can observe that DBTA dataset reflects real-world prevalence, resulting in a highly imbalanced distribution across classes. As for SRH imaging, we set 50\% of the known class WSIs as labelled during training.

\begin{table}[]
  \centering
  \begin{tabular}{|l|r|l|}
    \hline
    Tumor type & WSIs & known/novel \\ \hline
    Glioblastoma, IDH-wildtype                                    & 378 & known \\
    Pilocytic astrocytoma                                           & 138 &  novel \\
    Meningothelial meningioma                                       & 84 &  known\\
    Pituitary adenoma                                               & 79 &  known \\
    Ganglioglioma                                                    & 70 &  novel \\
    Grade 3, Oligodendroglioma, IDH-mut. and 1p/19q cod.    & 70 &  known\\
    Haemangioblastoma                                                & 69 &  known\\
    Grade 2, Oligodendroglioma, IDH-mut. and 1p/19q cod.               & 66 & known\\
    Atypical meningioma                                              & 66 &  novel\\
    Adamantinomatous craniopharyngioma                               & 66 &  novel\\
    Schwannoma                                                       & 65 &  known\\
    Diffuse astrocytoma, IDH-mut.                                  & 58 & novel\\
    \hline
  \end{tabular}
  \caption{Overview of the DBTA dataset for tumor types with more than 50 WSIs.}
  \label{tab:dbta}
\end{table}


\subsection{Implementation details}

All implementation were performed using Python (v3.10 Python Software Foundation) and PyTorch (v 2.5.1). The source code is avaiable at: \url{https://github.com/mperkonigg/HGCD_BT}.

\paragraph{SRH setting} As feature extractor $f$ we adopt the Vision Transformer (ViT-B-16) model \cite{dosovitskiy_image_2020} pre-trained on ImageNet using DINO \cite{caron_emerging_2021}, where we use the output \texttt{[CLS]} token as feature representations. Since the pre-training was performed on natural images, we choose to fine-tune the whole model during our training. The contrastive projection head $\phi$ is modeled as a four-layer multi-layer perceptron (MLP) with GeLu activation following the setup in \cite{vaze_generalized_2022,pu_dynamic_2023}. The projection head for hierarchical clustering $\pi$ is a two-layer MLP with ReLu activation after the first and sigmoid activation after the second layer with an output dimension of 7, corresponding to a tree level $T=3$, which is chosen based on the number of classes in the dataset. The generate different input views ($x_i$, $x_i'$) we applied data augmentation including horizontal and vertical flippling, randomized contrast adjustments, and Gaussian sharpening. We used a batch size of 64, a learning rate of 0.01 and trained for 200 epochs using stochastic gradient descent. For HGCD-BT we incorportated a 50-epoch warm-up phase before applying the hierarchical clustering loss. During this warm-up, we only use the contrastive loss (Eq. \ref{eq:contrastive_loss}) to adapt the pre-trained features of $f$ to the SRH imaging domain. Following \cite{vaze_generalized_2022} we set $\alpha=0.35$ in Equation \ref{eq:contrastive_loss}. We set $\beta=2^{-T}=2^{-3}=0.125$ in Equation \ref{eq:overall_loss}, following the recommendations in \cite{znalezniak_contrastive_2023}. To have equal contribution of contrastive learning and hierarchical clustering we set $\gamma=1$.

\paragraph{DBTA setting}
For DBTA we built on a domain-specific multimodel whole slide foundation model \textit{TITAN} as a slide-encoder \cite{ding_multimodal_2024} with a \textit{CONCH} patch encoder \cite{lu_visual-language_2024}. For processing we utilized the \textit{Trident} toolkit \cite{zhang_accelerating_2025}. Both \textit{TITAN} and \textit{CONCH} are using self-supervised visual-language alignment to train the models. \textit{TITAN} operates on a slide-level and is built on diverse large-scale set of WSIs spanning 20 different organs and including neoplastic and non-neoplastic tissue.  During training it extracts patch-level features (using \textit{CONCHv1.5}) that are aggregated to task-agnostic slide-level encodings. Due to the size and computational needs for inference and training on WSIs we extract slide encodings once prior to training. During GCD training we used a three layer MLP as feature encoder $f$ to transform those slide encodings to GCD embeddings. For contrastive learning the same projection head $\phi$ as for SRH patches is used. For hierarchical clustering the output dimension of $\pi$ is adapted to 31, corresponding to a tree level of $T=5$. Data augmentation is applied to the TITAN slide encodings, including Gaussian noise, scaling, shifting, and feature dropout, prior to transformation by $f$. A batch size of 16 was applied, the learning rate was set to 0.001 and the model was trained for 200 epochs using stochastic gradient descent. As for SRH we used a warm-up phase of 50 epochs where only the contrastive learning loss is applied. As for SRH we set $\alpha=0.35$ and $\gamma=1$. $\beta$ is adapted to the tree level of $T$ and set to $2^-5=0.03125$.

\subsubsection{Baselines}

We compare our approach to several baselines. We evaluate two state-of-the-art category discovery methods developed and tested for natural images GCD \cite{vaze_generalized_2022} and Dynamic Conceptional Contrastive Learning (DCCL) \cite{pu_dynamic_2023}. In addition, we use the features extracted from the pretrained models with DINO \cite{caron_emerging_2021} for SRH and TITAN for DBTA without any special attention to category discovery.
GCD \cite{vaze_generalized_2022} introduced the setting of generalized category discovery and proposed a simple semi-supervised contrastive learning approach. DCCL \cite{pu_dynamic_2023} is alternating between estimating visual conceptions, using a semi-supervised Infomap clustering, and learning representations during training. This setup allows DCCL to place classes (e.g. cats and dogs) from the same conception (e.g. animal) close together.

\subsubsection{Evaluation protocol}
As evaluation we follow the standard protocol in the GCD setting \cite{vaze_generalized_2022}. First, SemiKMeans clustering is performed on the predicted embeddings \cite{vaze_generalized_2022}. Next, the Hungarian assignment algorithm \cite{kuhn_hungarian_1955} is used to optimally align the predicted clusters and ground truth. Finally, accuracy scores are calculated on all unlabelled samples and reported for all classes, as well as separately known classes and novel classes.

\subsection{Results}

\subsubsection{Patch-level classification - OpenSRH}
Table \ref{tab:results_srh} shows a comparison of the accuracy of HGCD-BT and the baseline methods. The results show that the hierarchical approach of HGCD-BT is more suitable for patch-level brain tumor classification, as seen in the increase in performance between HGCD-BT and GCD (+38.2\%) and DCCL (+28.1\%). Especially for novel classes the difference between accuracies is high. Both GCD and DCCL show significant increase in performance compared to using pre-trained DINO features only. However, GCD is increasing the performance in novel categories the least (+7.0\%). Both DCCL and GCD are performing significantly better for tumor types partially labelled in the training set, whereas HGCD-BT shows a balanced, high accuracy.

\begin{table}[h!]
  \centering
  \begin{tabular}{|l|c|c|c|}
    \hline
    Method & All Acc. & Known Acc. & Novel Acc. \\
    \hline
    ViT (ImageNet pretrained DINO) & 35.7 & 35.5 & 36.0 \\
    GCD \cite{vaze_generalized_2022} &  55.5 & 70.9 & 43.0 \\
    DCCL \cite{pu_dynamic_2023} & 65.6 & 73.0 & 59.6 \\
    HGCD-BT (ours) & 93.7 & 93.1 & 94.2\\
    \hline
  \end{tabular}
  \caption{Results for HGCD-BT compared to baseline methods reporting overall accuracy and accuracies for known and novel classes separately.}
  \label{tab:results_srh}
\end{table}

\begin{figure}[h!]
  \centering
  \begin{subfigure}{0.45\textwidth}
    \includegraphics[width=\linewidth]{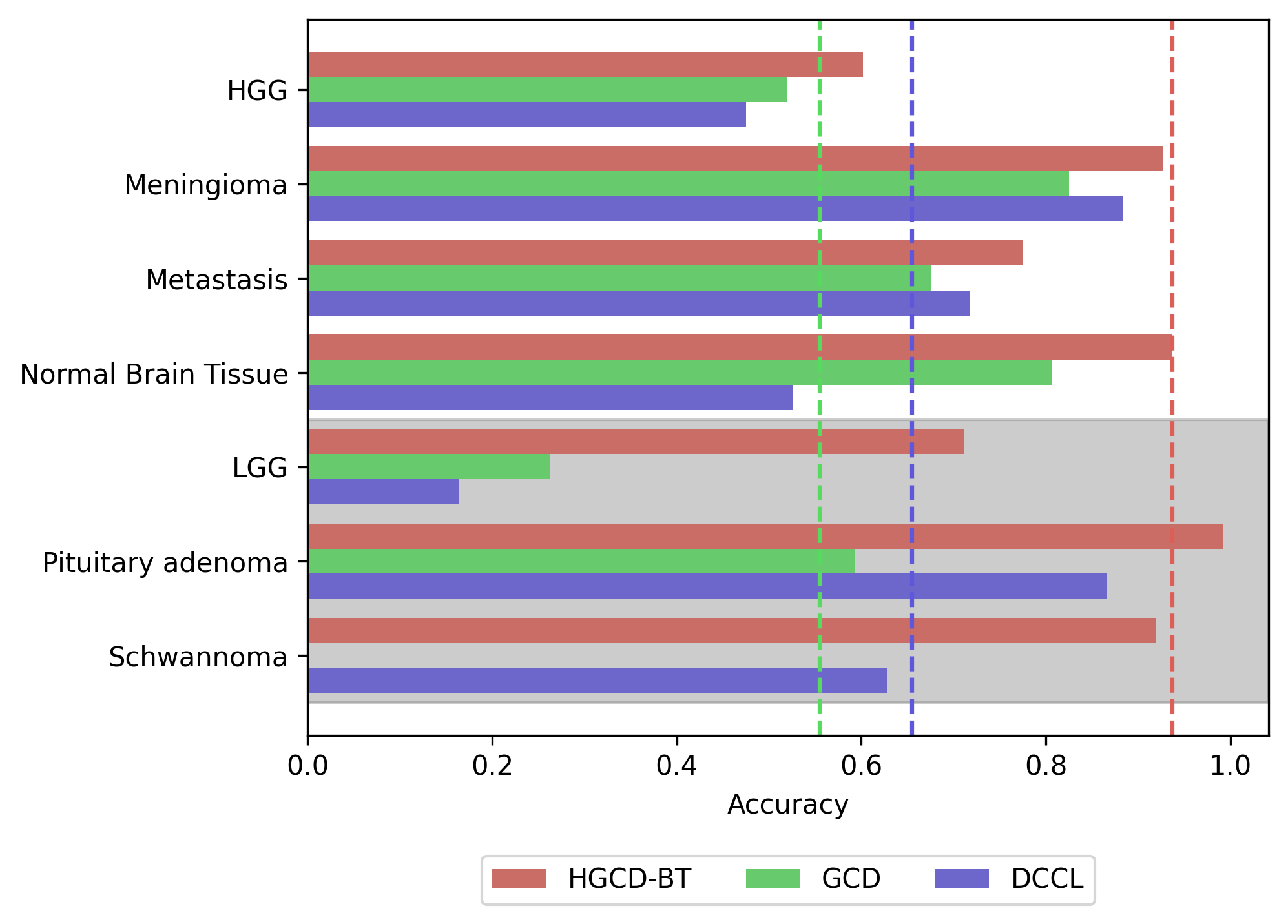}
    \caption{Class-wise accuracies of different types. Shaded region depicts novel classes, not labelled during training.}
    \label{fig:class_wise_srh}
  \end{subfigure}
  \hspace{0.05\textwidth}
  \begin{subfigure}{0.45\textwidth}
    \includegraphics[width=\linewidth]{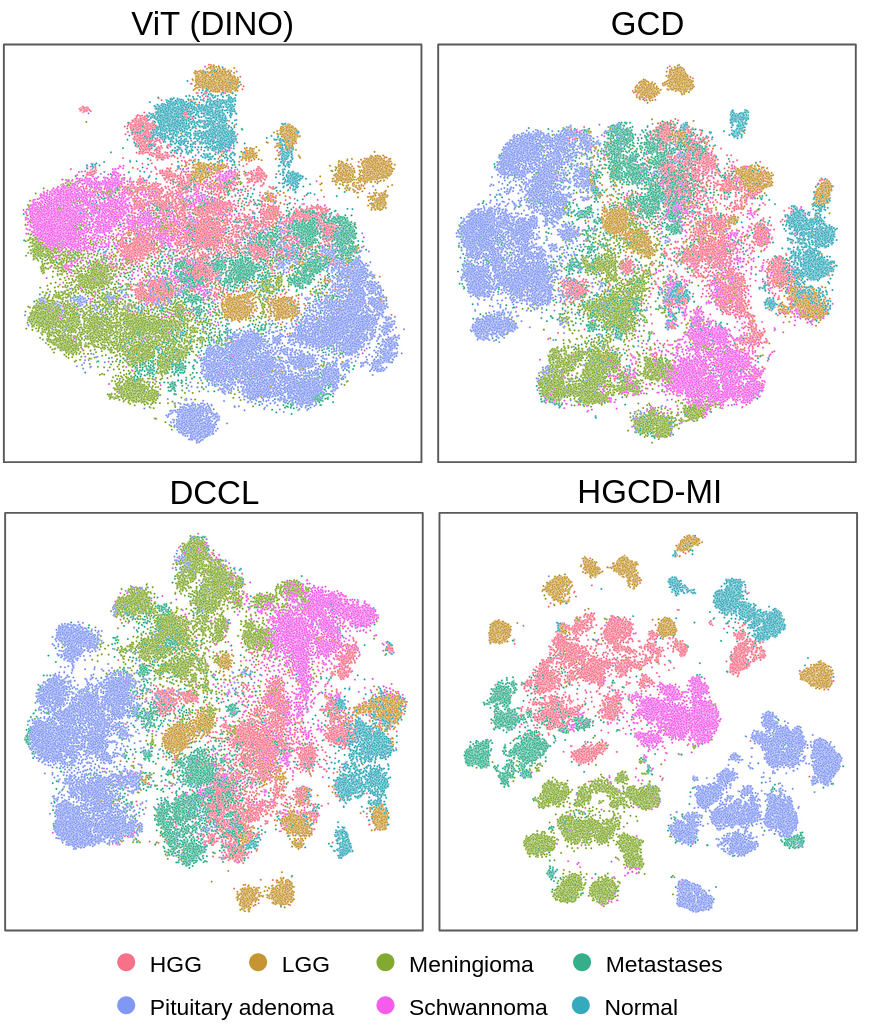}
    \caption{t-SNE \cite{maaten_visualizing_2008} visualization of feature embeddings.}
    \label{fig:tsne}
  \end{subfigure}
  \begin{subfigure}{\textwidth}
    \includegraphics[width=\linewidth]{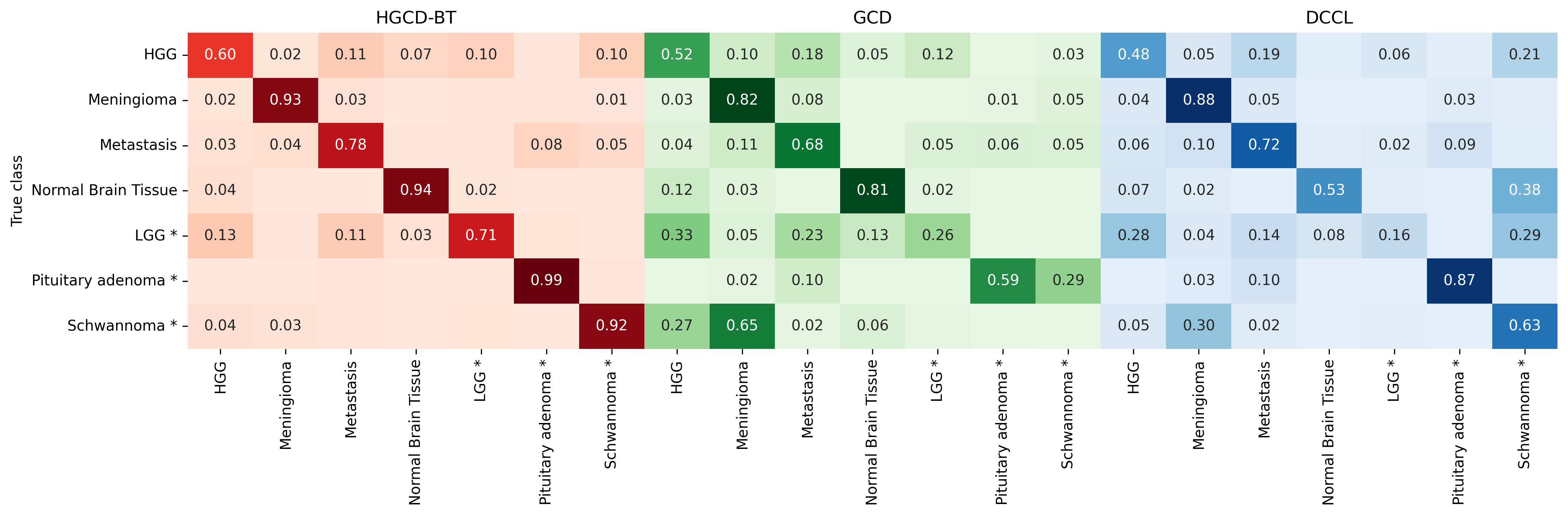}
    \caption{Confusion matrices between all classes, Novel classes are marked by *.}
    \label{fig:heatmap_srh}
  \end{subfigure}
  \caption{Results for GCD, DCCL and HGCD-BT for patch-level classification on OpenSRH.}
\end{figure}

We further analyse the performance differences by examining class-wise accuracies, as shown in Figure \ref{fig:class_wise_srh}. In particular, GCD fails to discover the category of schwannoma, a benign nerve sheath tumor, and it is mainly confused with meningionoma and HGG. HGCD-BT achieves significantly better accuracy for the novel brain tumor classes - LGG, pituitary adenoma, and schwannoma - compared to the other methods. Overall, HGG (in known classes) and LGG (in novel classes) exhibit lower accuracy compared to other categories. This is primarily due to high pairwise confusion between these classes and metastases. Biologically, this is expected, as both LGG and HGG belong to the glioma category, making their distinction particularly challenging. The hierarchical nature of the features extracted by the proposed method appears better suited for capturing subtle features such as increased cellularity, anaplasia and mitotic activity, differentating HGG and LGG.

Comparing the t-sne \cite{maaten_visualizing_2008} feature embedding visualization for the pre-trained ViT \cite{caron_emerging_2021}, GCD \cite{vaze_generalized_2022}, DCCL \cite{pu_dynamic_2023} and our HGCD-BT method in Figure \ref{fig:tsne} shows a better distinction of categories for HGCD-BT. While ViT exhibits a general trend toward separation, it lacks well-defined class boundaries. GCD and DCCL improve separation, but HGCD-BT achieves the most distinct class boundaries. Additionally, HGCD-BT reveals small, well-defined subgroups within categories, which can be further investigated.

\subsubsection{Slide-level classification - DBTA}
For WSI analysis the results in terms of accuracy are shown in Table \ref{tab:results_dbta}. Given that all compared methods are using a pathology foundation model as slide-level feature encoder the differences in accuracy is not as high as for SRH. Nevertheless, HGCD-BT reaches the best overall accuracy of 79.2\%, when compared to the pretrained TITAN feature encoder (+5.4\%), GCD (+7.2\%) and DCCL (+3.5\%). Especially for novel classes HGCD-BT shows strong performance (+8\%) against DCCL and the TITAN feature encoder.

\begin{table}[h!]
  \centering
  \begin{tabular}{|l|c|c|c|}
    \hline
    Method & All Acc. & Known Acc. & Novel Acc. \\
    \hline
    TITAN \cite{ding_multimodal_2024}/Conch\cite{lu_visual-language_2024} & 73.8 & 77.0 & 70.1 \\
    GCD \cite{vaze_generalized_2022} & 72.0  & 79.3  & 63.8 \\
    DCCL \cite{pu_dynamic_2023} & 75.7 & 80.6 & 70.1 \\
    HGCD-BT (ours) & 79.2 & 80.2 & 78.1\\
    \hline
  \end{tabular}
  \caption{Results for HGCD-BT compared to baseline methods reporting overall accuracy and accuracies for known and novel classes separately for DBTA.}
  \label{tab:results_dbta}
\end{table}

\begin{figure}[h!]
  \centering
  \begin{subfigure}{\textwidth}
    \includegraphics[width=\linewidth]{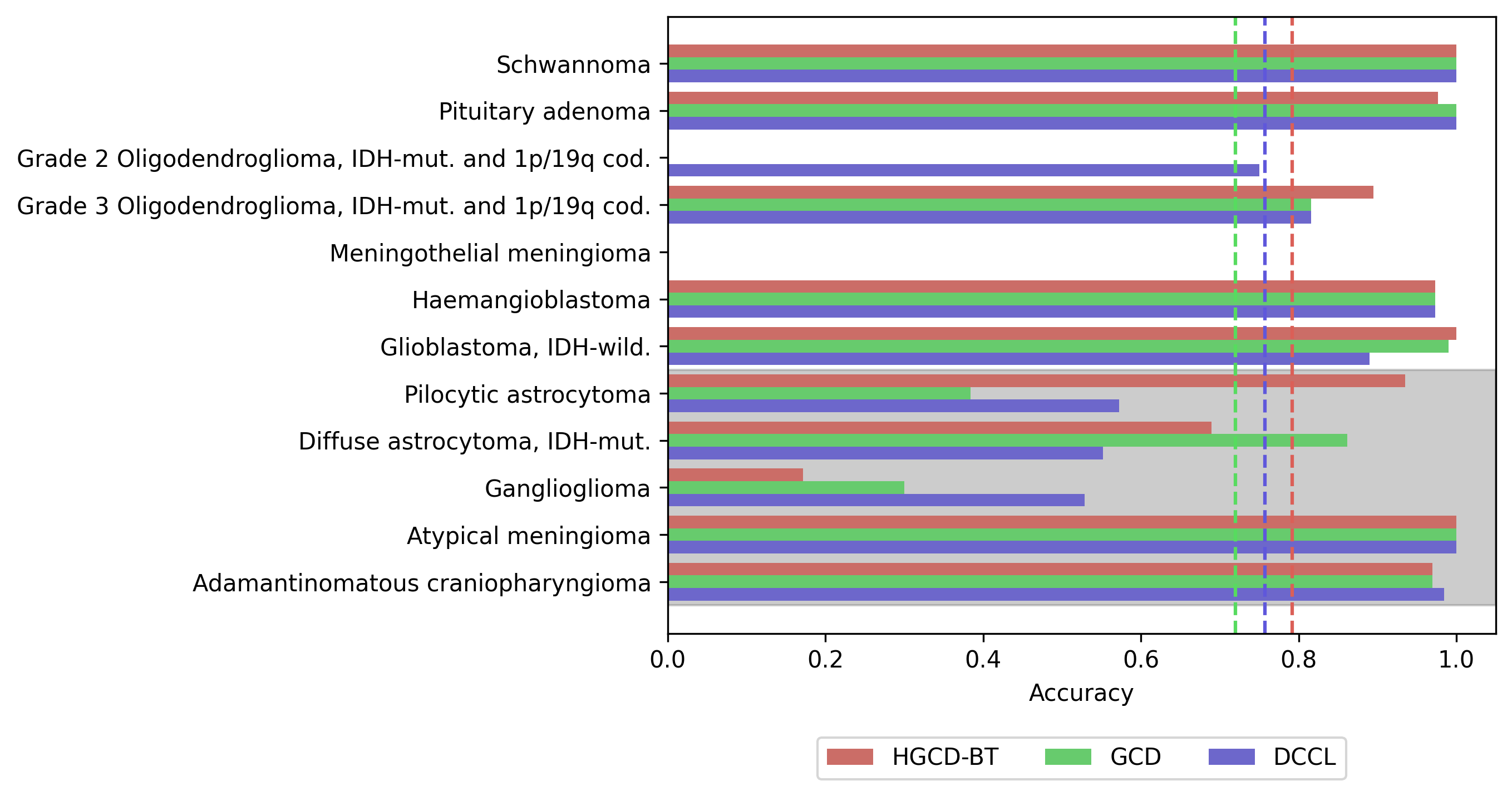}
    \caption{Class-wise accuracies of different types. Shaded region depicts novel classes, not labelled during training.}
    \label{fig:class_wise_dbta}
  \end{subfigure}
  \begin{subfigure}{\textwidth}
    \includegraphics[width=\linewidth]{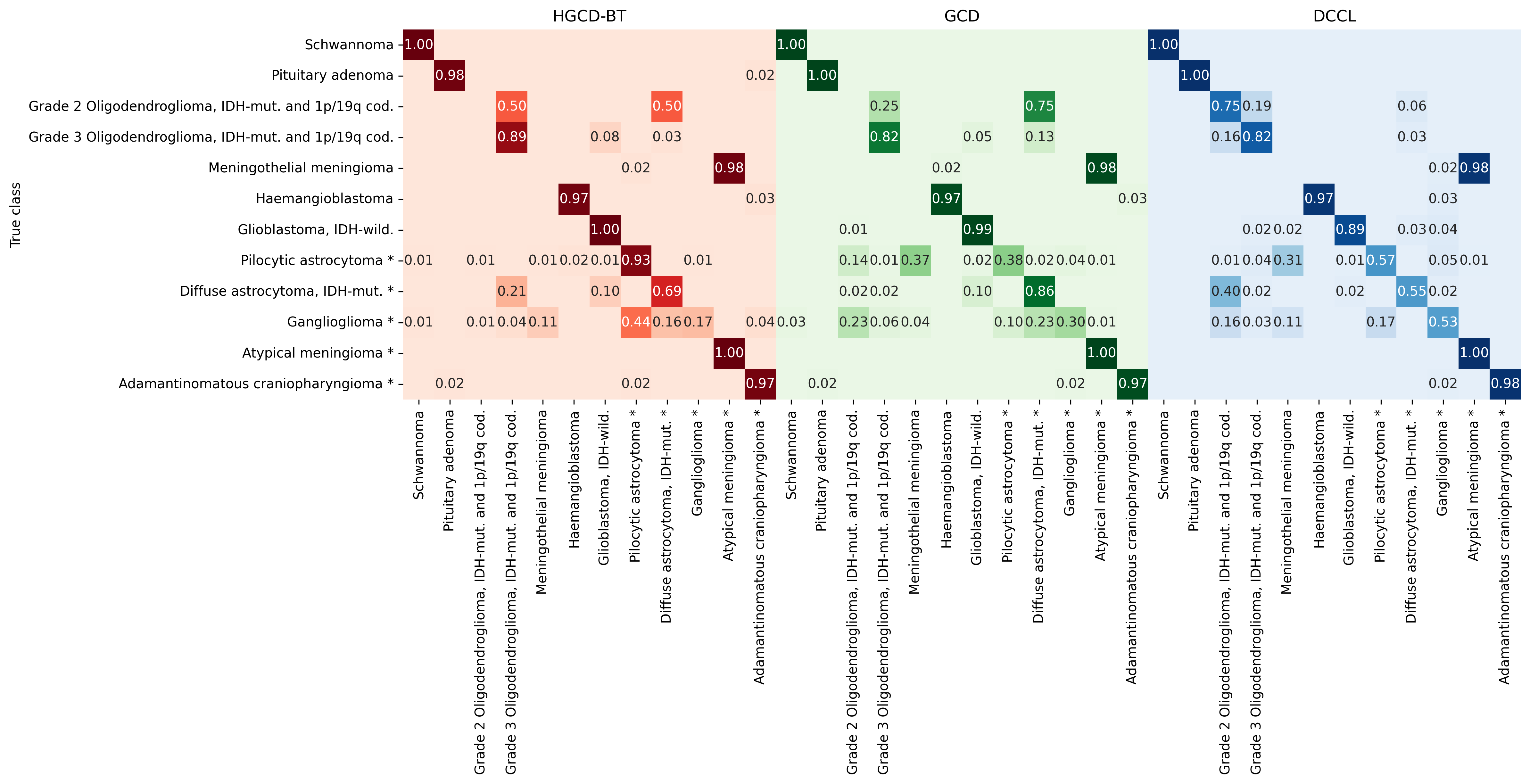}
    \caption{Confusion matrices between all classes, Novel classes are marked by *.}
    \label{fig:heatmaps_dbta}
  \end{subfigure}
  \caption{Results for GCD, DCCL and HGCD-BT for slide-level classification on DBTA.}
\end{figure}

For a more detailed analysis Figure \ref{fig:heatmaps_dbta} shows the performance for type-wise accuracy. We observe some tumor type confusions that are common across all methods and are explainable from a neuropathological standpoint. Meningothelial meningioma and atypical meningioma are merged into one class by all approaches, which is expected since both share cytological features and arrangements but differ mostly in mitotic frequency.
A frequent confusion, particularly for HGCD-BT and GCD, involves Grade 2 oligodendroglioma IDH-mut., 1p/19q cod., which is misclassified as either Grade 3 oligodendroglioma IDH-mut., 1p/19q cod. or diffuse astrocytoma. The distinction between Grade 2 and 3 oligodendroglioma is also visually difficult and lies along a morphological spectrum, therefore, it cannot be separated in the feature space. Grade 2 oligodendroglioma IDH-mut., 1p/19q cod. and diffuse astrocytoma both fall within the family of low-grade gliomas and  share histopathological features. HGCD-BT and GCD struggle to capture the fine-grained cues needed for accurate discrimination.
Finally, ganglioglioma is often confused with pilocytic and diffuse astrocytoma. Ganglioglioma is a mixed glio-neuronal tumour, comprised of a  glial and a ganglion component, where the glial component is often astrocytic.

\subsubsection{Ablation study}
We analyse the contributions of various elements of our proposed approach for patch-level on SRH imaging data.
\paragraph{Influence of key components}
Different key components of our method are studied in Table \ref{tab:ablation}. First, we can assess the importance of the inclusion of both the hierarchical loss $\mathcal{L}^h$ and the contrastive loss $\mathcal{L}^c$. Using only the components related to $\mathcal{L}^h$ (row 2) results in a drop of -2.3\% in overall accuracy. While using only $\mathcal{L}^c$ defaults to GCD \cite{vaze_generalized_2022}. Second, the novel extension of the hierarchical loss to a semi-supervised loss shows a increased performance (row 3 vs. row 5), especially for novel categories (+5.1\%).
Finally, the effect of warm-up phase used during training to adapt the features to the SRH imaging domain prior to performing hierarchical clustering is compared, while not using a warm-up phase before hierarchical clustering results in a drop of 5.3\% (9.2\% for novel classes) (row 4).

\begin{table}[]
  \centering
  \begin{tabular}{|l|cccc|c|c|c|}
    \hline
    & $\mathcal{L}^c$ & su. $\mathcal{L}^h$ & un. $\mathcal{L}^h$ & wu. & All Acc. & Known Acc. & Novel Acc. \\ \hline
    1 (GCD) \cite{vaze_generalized_2022} & \checkmark & - & - & - & 55.5 & 70.9 & 43.0 \\
    2 & - & \checkmark &  \checkmark & \checkmark &  91.4 & 90.1 & 92.0 \\
    3 & \checkmark & - & \checkmark & \checkmark &  90.1 & 92.8 & 89.1 \\
    4 &\checkmark & \checkmark & \checkmark & - &  88.4 & 92.8 & 85.0 \\
    5 (HGCD-BT) &\checkmark & \checkmark & \checkmark & \checkmark & 93.7 & 93.1 & 94.2\\
    \hline
  \end{tabular}
  \caption{Analysis of different components of our method. su. $\mathcal{L}^h$ / un. $\mathcal{L}^h$  = supervised and unsupervised hierarchical clustering loss respectively, wu. = warm-up phase}
  \label{tab:ablation}
\end{table}

\paragraph{Influence of tree levels $T$}

\begin{figure}[h]
  \centering
  \includegraphics[width=0.6\textwidth]{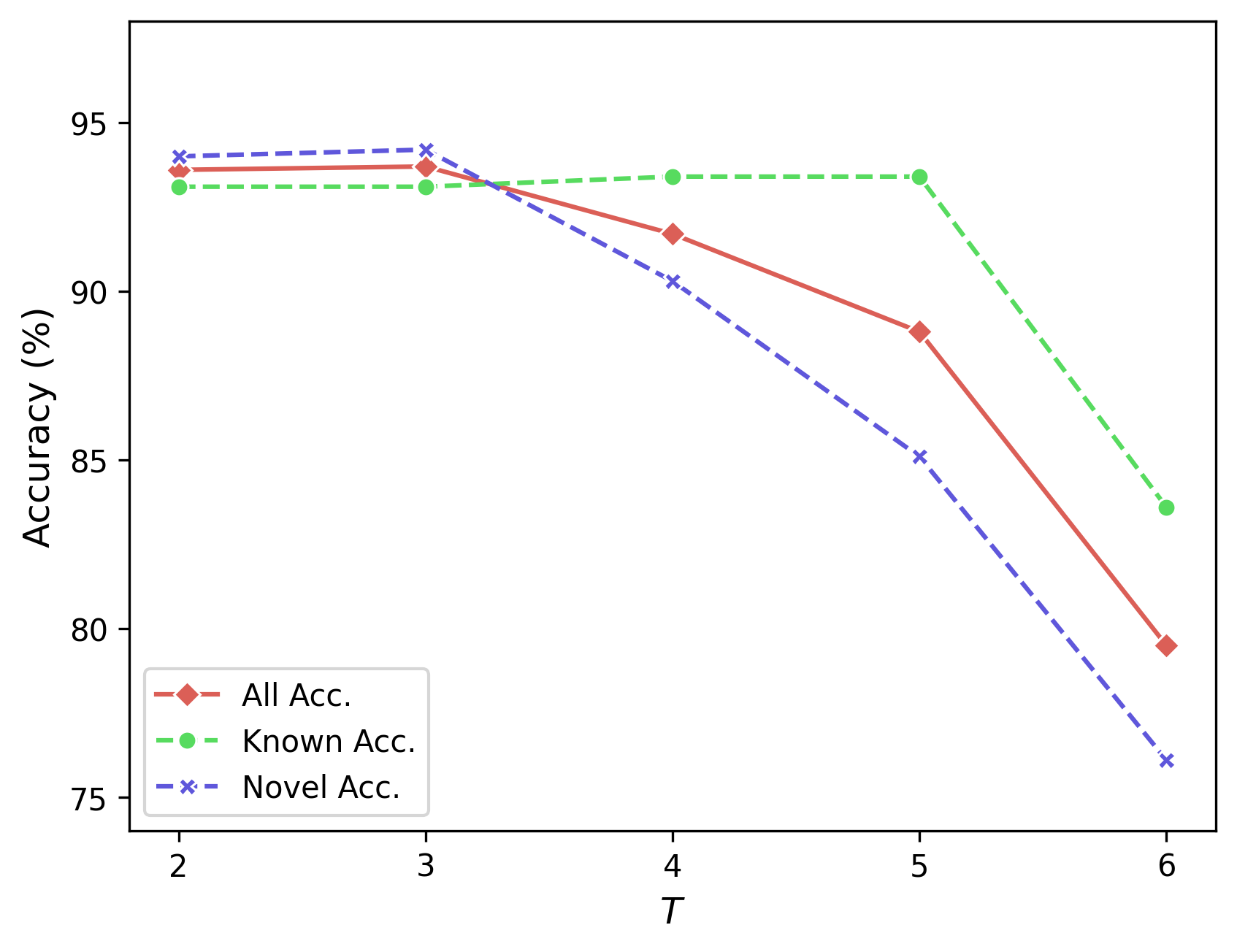}
  \caption{Different settings for the tree level $T$ for patch-level classification on OpenSRH}
  \label{fig:t_ablation}

\end{figure}

We test different values for the number of levels of the decision tree $T$ used during the calculation of $\mathcal{L}^h$ (see Figure \ref{fig:t_ablation}). For $T$ values between 2 and 4, we observe a stable overall accuracy ranging between 91.7\% ($T=4$) and 93.7\% ($T=3$). Adding more tree levels resulted in a performance loss, with $T=5$ accuracy on novel classes drops, while accuracy on known classes is stable. For an increase to $T=6$ the overall accuracy drops to 79.5\%, with both known and novel accuracy dropping significantly. This drop might be explained by a more challenging optimization process by adding more tree levels. In addition, tree level 3 results in eight leave nodes, close to the number of classes of seven which might support training. This observation is similar to the findings in \cite{znalezniak_contrastive_2023}.

\section{Discussion and Conclusion}
In this work, we introduced HGCD-BT, a novel approach that leverages hierarchical clustering and contrastive learning to improve category discovery in histopathological imaging. By incorporating a semi-supervised hierarchical clustering loss, our method effectively captures the structured nature of pathological patterns, outperforming existing GCD techniques. Through experiments on the OpenSRH and DBTA dataset, we demonstrated that HGCD-BT achieves superior classification performance, particularly in identifying novel classes not represented in the labelled part of the data. The results demonstrate generalizability across modalities (SRH and H\&E-staining) and level of classification (patch- and slide-level).
From a clinical perspective, more accurate and rapid identification of brain tumor types shortens the time of surgical interventions and may reduce complications and side-effects due to anesthesia and surgery. By discovering novel subtypes without requiring costly and labor-intensive annotations, HGCD-BT has the potential to support diagnostic workflows. In addition, using the hierarchical nature of HGCD-BT has the potential to improve interpretability of the decisions made by the model.
Our results demonstrate that concept discovery is a promising approach for brain tumor classification. The ability to uncover previously unrecognized categories could contribute to the development of more personalized treatment strategies, aligning with the broader goal of tailoring interventions to specific molecular or morphological subtypes. In addition, the newly identified classes might be incorporated in a specialized classification model using continual learning to enable a adaptive system without the need for retraining from scratch.
While we focused here on brain tumors, the methodology is broadly applicable to any disease domain with a hierarchical taxonomy. Potential applications include other tumor types (e.g., skin or thoracic cancers) as well as other disease areas such as lung infections, where hierarchical relationships naturally exist (e.g., lung infections → pneumonia → fungal pneumonia).

Despite the potential of HGCD-BT, there are a few limitations that need consideration. First, the number of tumor subtypes is restricted to the most common types. Data of rare tumor types is challenging to collect and fine-grained features are needed to distinguish the types. To fully realize its potential, future work should explore datasets containing rare diseases and address challenges related to long-tailed distributions.
In HGCGD-BT we need to set the number of $T$ manually, and our experiments confirm that performance is sensitive to this choice. Future work should explore strategies how to estimate $T$ dynamically to allow for a flexible extension of the classes represented in the data. Similarly, for evaluation we use the standard approach for GCD evaluation, which relies on a semi-supervised version of k-means clustering. This requires setting the number of clusters. While this enables direct comparision to other methods, future work should explore the use of the learned hierarchy in the hierarchical clustering branch directly to make a classification.
To construct the tree hierarchy, we are following \cite{znalezniak_contrastive_2023} in constructing a binary tree to facilitate training. Nevertheless, this restricts the flexibility of the approach, hierarchies for brain tumor subtypes do not follow a strictly binary decision tree. Extending HGCD-BT to allow non-binary hierarchies would more accurately reflect true pathological relationships.
Finally, our work considered imaging data alone, extending the method to multimodal data, by integrating genomic, radiological, or clinical data, could offer new opportunities for discovering disease trajectories and a more comprehesive modelling of diseases.

\section*{Acknowledgments}
The computational results have been achieved in part using the Austrian Scientific Computing (ASC) infrastructure.

\bibliographystyle{splncs04}
\bibliography{references}

\end{document}